\documentclass[10pt, ,letterpaper]{article}
\usepackage[top=0.85in,left=.8in,footskip=0.75in,marginparwidth=2in]{geometry}

\usepackage{cite}
\usepackage{amsmath}
\usepackage{lineno}
\usepackage{nameref,hyperref}

\usepackage{microtype}
\DisableLigatures[f]{encoding = *, family = * }

\usepackage{changepage}
\usepackage[aboveskip=1pt,labelfont=bf,labelsep=period,singlelinecheck=off]{caption}

\makeatletter
\renewcommand{\@biblabel}[1]{\quad#1.}
\makeatother

\usepackage{soul}
\usepackage{graphicx}
\usepackage{hyperref}
\usepackage{placeins}
\usepackage{mathptmx}
\usepackage{anyfontsize}
\usepackage{t1enc}
\usepackage{lastpage,fancyhdr,graphicx}
\usepackage{epstopdf}
\usepackage{cite} 
\usepackage{multirow}
\usepackage{url}
\usepackage{tabularx}
\usepackage{placeins}
\usepackage{multicol,lipsum}
\usepackage{rotating}
\usepackage{natbib}
\fancyheadoffset[L]{2.25in}
\fancyfootoffset[L]{2.25in}

\usepackage{color}

\definecolor{Gray}{gray}{.25}

\usepackage{graphicx}

\usepackage{sidecap}

\usepackage{wrapfig}
\usepackage[pscoord]{eso-pic}
\usepackage[fulladjust]{marginnote}
\reversemarginpar

\begin{document}
	\vspace*{0.35in}
	
	\begin{flushleft}
		{\Large
			\textbf\newline{QLMC-HD: Quasi Large Margin Classifier based on Hyperdisk}
		}
		\newline
		\\
		Hassan Ataeian\textsuperscript{1},
		Shahriar Esmaeili\textsuperscript{2},
		Saeideh Roshanfekr\textsuperscript{3},
		Neda Maleki Khas\textsuperscript{4},
		Ali Amiri\textsuperscript{1}, and
		Hossein Safari\textsuperscript{5}
		\\
		\bigskip
		\bf{1} Department of Computer Engineering, University of Zanjan, University Blvd., Zanjan, 45371-38791, Iran\\
		\bf{2} Department of Physics and Astronomy, Texas A$\&$M University, 4242 TAMU, University Dr., College Station, TX 77843, US\\
		\bf{3} Department of Computer Engineering and Information Technology, Amirkabir University of Technology, 424 Hafez Avenue, 15875-4413 Tehran, Iran\\
		\bf{4} Department of Biology, Islamic Azad University - Zanjan Branch, Zanjan, 58145-45156, Iran\\
		\bf{5} Department of Physics, University of Zanjan, University Blvd., Zanjan, 45371-38791, Iran
		\bigskip

	\end{flushleft}
	\justify
	\section*{Abstract}
	In the area of data classification, the different
	classifiers have been developed by their own strengths and weaknesses. Among these classifiers, we propose a method that is based on the maximum margin between two classes. One of the main challenges in this area is dealt with noisy data. In this paper, our aim is to optimize the method of large margin classifiers based on hyperdisk (LMC-HD) and combine it into a quasi-support vector data description (QSVDD) method. In the proposed method, the bounding hypersphere is calculated based on the  QSVDD method. So our convex class model is more robust compared with the support vector machine (SVM) and less tight than LMC-HD. Large margin classifiers aim to maximize the margin and minimizing the risk.  {Since} our proposed method ignores the effect of outliers and noises, so this method has the widest margin compared with other large margin classifiers. In the end, we compare our proposed method with other popular large margin classifiers by the experiments on a set of standard data which indicates our results are more efficient than the others.
	
	\textbf{Key words:} Classification, Hyperdisk, Large margin classifier, Kernel method, Support Vector Machine

	\section{Introduction}\label{sec1}
	
	In the past years, classification based on large margin classifier has been one of the topics taken into consideration by researchers in machine learning \cite{Parikh_2016, Panwar_2012, Wu_2013, Alabdulmohsin_2016, Campbell2000QueryLW, Shen:2015:SLM:2888116.2888129, Gupta_2018, Cho2018LargeMarginCI, Libbrecht_2015, Leung_2016, Moradkhani_2015}. Support vector machine (SVM) is one of the most common methods among these classifiers, which is derived from statistical learning theory and has created great success in pattern recognition \cite{Carvalho_2017, Mar_nov__2017}. 
	
	SVM was introduced by Cortes and Vapnik in 1995 which originated on the basis of statistical theory and structural risk minimization \cite{Cortes_1995}. Since SVM seeks to create the largest margin between two-class data, it is considered as large margin methods and is part of kernel methods in machine learning \cite{tan2006introduction}.  The main motivation of the method is to find the hyperplane that separates the data of these two classes with maximum margins. In this method, the data of each class will be modeled with a geometric model called the convex hull. 
	
	The convex hull, due to its tight structure toward each class attributes, creates a more closed model of each class data; that is, it covers a more restricted area than the real area of each class data. On the other hand, the support vector machine is designed for the state where the convex hull of each class data does not have much overlapping in the space of data feature (or space where the kernel is applied to the data). Also, if the data convex hull does not have overlapping, yet the support vector machine will have the former problem and cannot correctly identify the true extent of data for each class. To solve this problem regarding convex hull, other geometric models, such as hypersphere and affine hull, were introduced by \cite{Cevikalp_2009, Cevikalp_2010, Cevikalp_2013}.

	Large margin approaches are created based on geometrical objects like hypersphere, convex hull, affine hull, and hyperdisk. The classification based on large margin methods has been progressing remarkably in computer vision, text analysis, biometrics, and biometrics rather than previous methods \cite{Cevikalp_2010, Gupta_2018}. Actually, these methods are initially modeled each classes data using one of those geometrical methods and identifying the class of data, then treat with one of the following approaches.
	
	The first classification approach is based on the test data distance to the geometric model of each class data. In this method, the test data belongs to the class with the closest distance to the geometric model of that class. Nearest Convex Hull (NCH), Nearest Affine Hull (NAH), and Nearest Hyper Disk (NHD) classifications  determine the test data based on this approach. 
	
	In the second approach, a maximum margin will be determined between geometric models of each class data. The line existing in the middle of this margin and parallel to the margin will be identified as decision boundary and separating line, and classification will be done based on the fact the test data are located on which side of the line. SVM, large margin classifier based on affine hulls (LMC-AH), and LMC-HD classifications determine the test data based on the second approach. Adopting the second approach in determining test data class provides a better result at run-time \cite{Cevikalp_2013, Allahyari2012QuasiSV, Han_2012, JMLR:v15:delgado14a, chen1984linear, kb2005, Cevikalp_2009}.

	It is mentioned in the literature that the hyperspheres have a good performance in detecting the noises.  Also, the classifications based on affine hulls work good in some cases but not in all cases, although in total they work better than classifications based on the convex hull.  Obviously, the hyperdisk geometric model in an area related to single area data represents a smaller area than an affine hull for the same data, which is similar to the support vector machine in this respect. Combining affine hull and hypersphere leads to having both methods advantages and minimizing the disadvantage of each one to some extent \cite{Cevikalp_2013, Cevikalp_2009}.

	The disadvantage of SVDD method is that it does not take into account the gravity center of data, and the center and radius of the hypersphere will be estimated farther away from its actual radius and center. To solve this problem,  \cite{Allahyari2012QuasiSV} proposed QSVDD method that involves the optimization of the gravity center of data in target function. So, the impact of the noises on the center and the radius of hypersphere is reduced, and the estimate of its center and radius will be closer to the reality. The problem of SVDD method in dealing with the noises on hyperdisk, obtained from the intersection between data affine hull and the hypersphere inscribed on the data, is also evident in LMC-HD method. Therefore, to solve this problem, quasi-large margin classifier based on hyperdisk method (QLMC-HD) is proposed.

	This paper is organized as follows, our proposed method (QLMC-HD) is introduced in Section \ref{sec2}. In Section \ref{sec3}, the results of implementing our proposed method are presented. To test the robustness of the proposed method, different types of data set are tested. The main conclusions is also presented in Section \ref{sec4}.

	\section{Method (QLMC-HD)}\label{sec2}

	 {This paper presents a quasi-large margin classifier based on the hyperdisk method (QLMC-HD) for reducing the impact of the noises on the center of the hypersphere. This method has improved the LMC method and indicates a better margin with the gravity center. The gravity center is the center of data based on density, and it means that the gravity center is the mean of the dataset without noise and outlier.} The data will be modeled with hyperdisk in LMC-HD method. Hyperdisk is the intersection of the affine support data and the smallest bounding hypersphere inscribed on the training samples in feature space. The center of the hypersphere will be determined according to the support vectors and is influenced by them. On the other hand, in support vector data description (SVDD) method, limiting Lagrange coefficients cannot have much effect on the role of noisy data set and outliers, and the center and radius of the hypersphere yet will be affected by these data \cite{Allahyari2012QuasiSV}. So, based on the above argument on the failure of SVDD method to deal with outliers and the negative impact of outliers on the hypersphere center, this problem also exists in hyperdisks, and in fact, the radius and center of hyperdisks in LMC-HD method are also misestimated under the influence of outliers. To solve SVDD problem, QSVDD method is introduced. In this method, to reduce the impact of the outliers on the center and radius of the hypersphere, the gravity center of data is taken into consideration and is combined effectively with SVDD method.
	
\begin{figure*}[ht]
		\centerline{\includegraphics[width=1.1\textwidth,clip=]{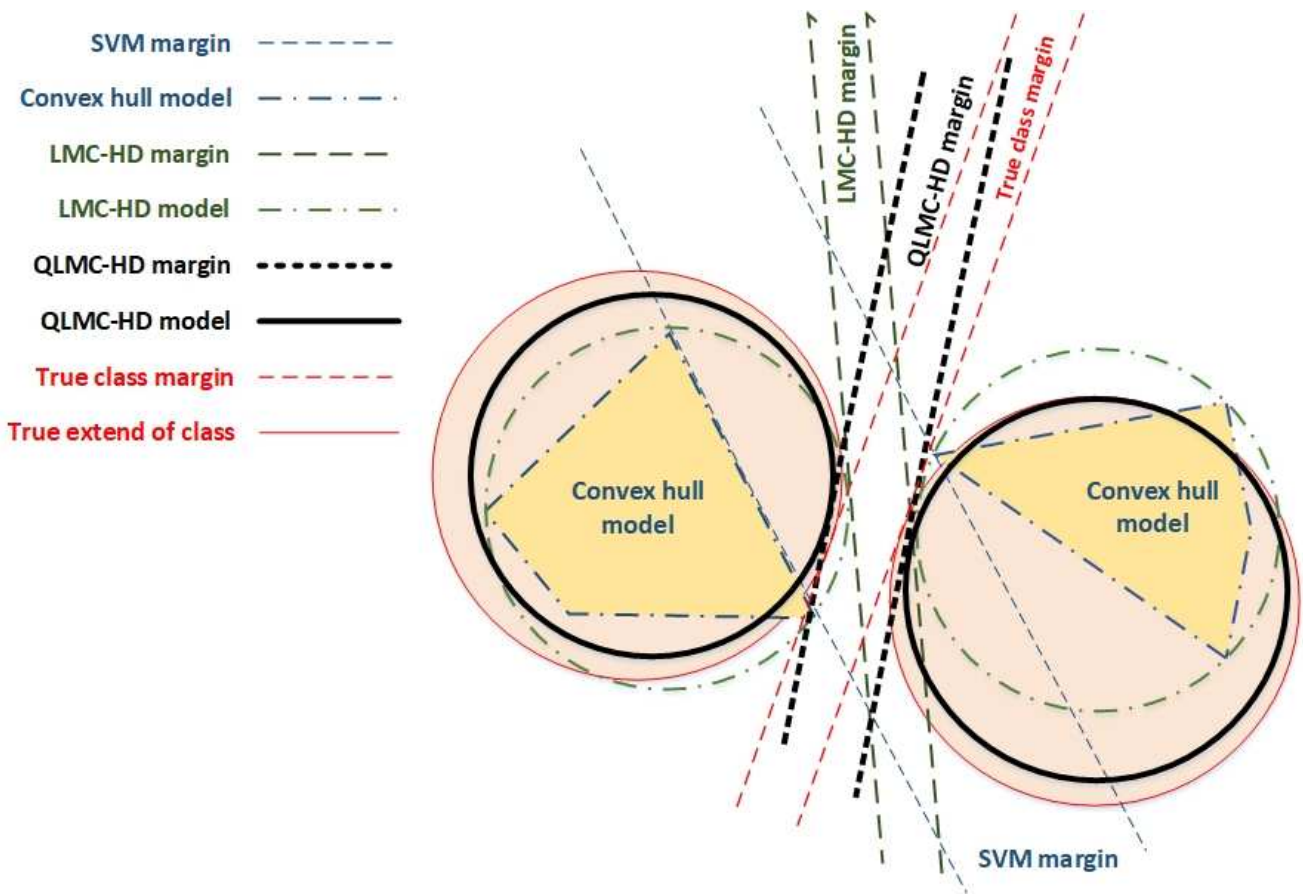}}
		\caption{ {Schematic of QLMC-HD model and other methods; Red line: True extend of classes; Dashed red line: True class margin; Dotted dashed blue line: Convex hull of SVM model; Dashed blue line: SVM margin;  Dotted dashed green line: LMC-HD model; Dashed green line: LMC-HD margine; Bold black line: QLMC-HD model; Dashed black line: QLMC-HD margin.}}\label{fig1}
	\end{figure*}

	Our idea in this article is to consider the gravity center of data at the time of constructing hyperdisk in LMC-HD method and to introduce the intersection between affine hull and hypersphere constructed based on QSVDD method as a new technique for modeling the data. Therefore, it is obvious that because hyperdisk is the result of an intersection between affine hull and hypersphere data, and because the geometric model of affine hull is a larger set, the intersection between this set and the optimized geometric model of hypersphere forms a model that will have a better performance in dealing with outliers than hyperdisk geometric model in LMC-HD method. We called this method QLMC-HD, due to its intense similarity with LMC-HD.  {Considering} the influence of the gravity center of data during hyperdisk formation is employed in QLMC-HD. Figure \ref{fig1} illustrates the schematics of SVM, LMC-HD, and QLMC-HD model.

	In the following, the method of finding the intersection between the affine hull and QSVDD-based hypersphere will be described mathematically in details. According to our idea to determine the center $(a)$ and radius  $(R)$ of the sphere in hyperdisk model, the equation \ref{eq1}
	\begin{equation}\label{eq1}
		H^{disk}_{c}=\{X=\sum^{n_{c}}_{i=1} \alpha_{i}x_{ci}\Big | \sum^{n_{c}}_{i=1} \alpha_{i} = 1, ||x-a||\leq R\}
	\end{equation}
	can be derived where where $a$ is the center of the hyperdisk, $R$ is its radius, $n_{c}$ is the number of instance of class $c$ and   {$\alpha$ is a coefficient. The objects with the coefficients $\alpha_{i}>0$ are called the support vectors.} Equation \ref{eq1} is actually all the points related to  {affine} hull considering the constraint $||x-s_{c}||\leq r_{c}$  \cite{Tax_1999, Cevikalp_2013}. If we call the center and radius of the hypersphere in this new method $a^{new}$ and $R^{new}$, then according to QSVDD method, the distance between sphere center and the gravity center of data will be as follows,
	\begin{equation}\label{eq2}
		|\Bar{x}-a^{new}|^2=|\sum^n_{i=1}\frac{x_{i}}{n} -a^{new}|^2 = \frac{1}{n^2}|\sum^n_{i=1} x_{i} - \sum^n_{i=1} a^{new}|^2 = \frac{1}{n^2} \sum^n_{i=1}\sum^n_{j=1} (x_{i}-a^{new})^T(x_{j}-a^{new}).
	\end{equation}
	Now the problem can be modeled as follows,
	\begin{equation}\label{eq3}
		\begin{matrix}
			\begin{matrix}
				min \\
				
			\end{matrix} & (R^{new})^2+\frac{B}{n^2}  \sum^n_{i=1}\sum^n_{j=1}(x_{i}-a^{new})^T(x_{j}-a^{new})+c\sum^n_{i=1}\epsilon_{i}  \\
			& \\
			subject~to & (x_{i}-a^{new})^T(x_{i}-a^{new}) \leq (R^{new})^2 +\epsilon_{i},\quad \epsilon_{i}\geq 0, \quad i=1,2,…,n.
		\end{matrix}
	\end{equation}
	In the above statement, parameter $B$ shows the extent of giving importance to the gravity center of data, and the rest of the parameters are similar to the $SVDD$ state \cite{Tax_1999}.  {The parameter $\epsilon$ is a slack variable, $C$ is a regularization parameter that controls the trade-off between maximizing the margin and minimizing the training error. Small $C$ tends to emphasize the margin while ignoring the outliers in the training data, while large $C$ may tend to over fit the training data.} Now, Lagrange equations will be formed as,
	\begin{equation}\label{eq4}
		L=(R^{new})^2+\frac{B}{n^2}  \sum^n_{i=1}\sum^n_{j=1}(x_{i}-a^{new})^T(x_{j}-a^{new})+c\sum^n_{i=1}\epsilon_{i}+c\sum^n_{i=1} \alpha_{i}((x_{i}-a^{new})^T(x_{i}-a^{new}) - R^2 - \epsilon_{i}) - \sum^n_{i=1} \epsilon_{i}\mu_{i}
	\end{equation}
	Now, if we differentiate the above equation in terms of  $a^{new}$, $R^{new}$ and $\epsilon_{i}$ and replace the resulted statement in Lagrange equations, 
	\begin{equation}
		\frac{\partial L}{\partial R^{new}} = 0,\quad -2R^{new}-2R^{new}\sum^n_{i=1} \alpha_{i} = 0 \quad \rightarrow \quad \sum^n_{i=1} \alpha_{i} =1,
	\end{equation}
	\begin{equation}
		\frac{\partial L}{\partial \epsilon_{i}} = 0,\quad \alpha_{i}+\mu_{i} = c \quad \rightarrow \quad 0\leq \alpha_{i}\leq c,
	\end{equation}
	\begin{equation}
		\frac{\partial L}{\partial a^{new}} = 0,\quad \rightarrow \quad \frac{B}{n^2}\sum^n_{i=1}\sum^n_{j=1}(-(x_{i}-a^{new})-(x_{j}-a^{new}))\sum^n_{i=1}\alpha_{i}(-2(x_{i}-a^{new})) = 0 \quad \rightarrow \quad a^{new} = \frac{\sum^n_{j=1} (\alpha_{i}+\frac{B}{n})x_{i}}{1+B}
	\end{equation}
	then $L_{D}$  {(Lagrangian dual)} will be obtained. Since $L_{D}$ is in terms of $\alpha_{i}$, then the statement should be maximized as,
	\begin{equation}\label{eq5}
		\begin{matrix}
			\begin{matrix}
				max \\
				
			\end{matrix} & \frac{1}{(1+B)}\left(-2\frac{B}{n^2}\sum^n_{i=1}\sum^n_{j=1}\alpha_{i} x_{i}^{T} x_{j} - \frac{B}{n^2}\sum^n_{i=1}\sum^n_{j=1} x_{i}^{T} x_{j} - \sum^n_{i=1}\sum^n_{j=1}\alpha_{i} \alpha_{j} x_{i}^{T} x_{j}\right) + \sum^n_{i=1}\alpha_{i} x_{i}^{T} x_{j} + \frac{B^2}{n^2}\sum^n_{i=1}\sum^n_{j=1} x_{i}^{T} x_{j}\\
			& \\
			subject~to & \sum^n_{i=1} \alpha_{i} = 1, \quad 0\leq \alpha_{i} \leq c, \quad i=1, 2,..., n, \quad -1<B<+\infty.
		\end{matrix}
	\end{equation}
	 {This optimization problem is solved for each of class.} By solving this second order programming problem, $\alpha_{i}$ will be achieved and then the center of hypersphere will be calculated from 
	\begin{equation}\label{eq6}
		a^{new}=\frac{\sum^n_{i=1}(\alpha_{i}+\frac{B}{n})x_{i}}{1+B}.
	\end{equation}
	In this regard, the instance $z$ will be inside the hypersphere if 
	\begin{equation}\label{eq7}
		|z-a^{new}|^2 = z^{t}z-\frac{2}{(1+B)}\left(\sum^n_{i=1}\alpha_{i} x_{i}^{T} z + \frac{B}{n}\sum^n_{i=1} x_{i}^{T}z\right)+\frac{1}{(1+B)^2}\left(\sum^n_{i=1}\sum^n_{j=1}\alpha_{i} \alpha_{j} x_{i}^{T} x_{j}+ \frac{B^2}{n^2} \sum^n_{i=1}\sum^n_{j=1} x_{i}^{T} x_{j}+2\frac{B}{n} \sum^n_{i=1}\sum^n_{j=1}\alpha_{i} x_{i}^{T} x_{j}\right) \leq (R^{new})^2.
	\end{equation}
	
	Now, by calculating the center and radius of the sphere, to find the maximum margin between hyperdisks and their separating line, it is enough to find the closest point of each hyperdisk and find the perpendicular line passing from these points. If the closest points be $x_{-}$, $x_{+}$, then $<w,x>+b=0$ will be the line separating two hyperdisk with maximum margin, wherein this equation $w=\frac{(x_{+}- x_{-})}{|(x_{+}-x_{-})|}$ and $b=-<w,\frac{(x_{+}+ x_{-})}{2}>$. 
	Now, with regard to hyperdisk model, there are two ways to find two points close to each other that are similar to $LMC-HD$ method, with this difference that the center and radius of the hyperspheres are different. In this research, $QCQP$ method is used to calculate two close points on hyperdisk. $QCQP$ optimization problems refer to the issues where both objective function and constraints are second ordered. 
	Suppose that $X_{+}$  and $X_{-}$  are matrices that their columns equal negative and positive class data of education set, respectively. In this case, if two points $x_{\pm}$ be the close points on hyperdisk, $x_{\pm}$ can be defined so that the $x_{\pm}=X_{\pm}\alpha_{\pm}$ equation holds true. Now, the problem will be modeled as follows,

		\begin{equation}\label{eq8}
		\begin{matrix} 
			\begin{matrix}
				min \\
				
			\end{matrix} & |X_{+}\alpha_{+} - X_{-}\alpha_{-}|^2  \\
			& \\
			subject~to & \sum^{n_{+}}_{i=1} \alpha_{+i}=1, \quad \sum^{n_{-}}_{j=1} \alpha_{-j}=1, \quad |X_{+}\alpha_{+} - \alpha^{new}_{+}|^2 \leq (R^{new})^2_{+},\quad |X_{-}\alpha_{-} - \alpha^{new}_{-}|^2 \leq (R^{new})^2_{-}
		\end{matrix}
	\end{equation}

	$x_{\pm}$ can be calculated by obtaining $\alpha_{\pm}$ by solving the above problem. Accordingly, the hyperplane separating the data can be determined.

	$QCQP$ problems can be converted to Semi-definite Programs (SDP) problems. $CVX$\footnote{http://cvxr.com/cvx.} toolbox uses this idea, but in our implementations, this method sometimes did not lead to an answer or produced the wrong answers.  {Wrong answer means the answer that is not optimal. In generally, CVX toolbox sometimes failed to find a solution or returned an incorrect one.} So, in this research, $MOSEK$\footnote{http://www.mosek.com} toolbox is used to solve this problem. $MOSEK$ toolbox usually converts $QCQP$ problems to Second-Order Cone Program (SOCP) problems, which in polynomial time can be solved by Interior point method and has a performance better than the SDP method. The above problem easily can be solved by $MOSEK$ toolbox.  When there is an overlap between two class hyperdisks, two solutions can be suggested. The first solution is to constraint $c$ parameter in equation \ref{eq5} and assign a number less than 1, so that a more compact hypersphere result, and therefore, the outliers that have been the reason of hyperdisks overlap will be ignored. In case the data totally be non-linear and the data set is too complex, Kernel trick can be used. So, in $QCQP$ problem, due to the existence of interior multiplication, Kernel trick is used.

	Suppose we have $k(x_{i},x_{j})=<\phi(x_{i}),\phi_(x_{j})>$ as Kernel function and $\phi_{+}=[\phi(x_{1}^{+}),...,\phi(x_{n_{+}}^{+})]$ and $\phi_{-}=[\phi(x_{1}^{-}),...,\phi(x_{n_{-}}^{-})]$ are considered as kernel data matrices. We define 
	\begin{equation*}
		\left(
		\begin{matrix}
			
			k_{+} & -k_{+-}\\
			k_{+}^{T} & k_{-}
			
		\end{matrix}
		\right),
	\end{equation*}
	 {$\beta_{+}$ and $\beta_{-}$ are hypersphere centres in positive and negative classes} and the center of the positive and negative class hypersphere that map in feature space will be considered $k_{+}\beta_{+}$  and $k_{-}\beta_{-}$, respectively. In this case, the Kernel ed $QCQP$ problem will be as follows
	\begin{equation}\label{eq9}
		\begin{matrix} 
			\begin{matrix}
				min \\
				
			\end{matrix} & \alpha^T Ka  \\
			& \\
			subject~to & \alpha_{+}^T e_{+}=1, \quad \alpha_{-}^T e_{-}=1, \quad \alpha_{+}^T K_{+} \alpha_{+}-2\alpha_{+}^T K_{+} \beta_{+}+\beta_{+}^T K_{+} \beta_{+} \leq (R^{new})_{+}^2 		
			,\quad \alpha_{-}^T K_{-} \alpha_{-}-2\alpha_{-}^T K_{-} \beta_{-}+\beta_{-}^T K_{-} \beta_{-}  \leq (R^{new})_{-}^2
			
		\end{matrix}
	\end{equation}
	
	Applying this kernel leads to an increase in the speed and a decrease in calculations.  {$\beta_{+}$ and $\beta_{-}$ can be used in order to decreas the amount of computation required. It can also be considered as the centers which located on the sphere and depend only on the training point.} By finding wheits of $\alpha$, the  classifier  $\alpha_{+}^T K_{+x}-\alpha_{-}^T K_{-x}+b>0$ will be calculated. In this equation, $K_{+x}$ and $K_{-x}$ are Kernel vectors of the test sample against positive and negative data, which in this unequal
	
	\begin{equation*}
		b = \frac{-(\alpha_{+}^T K_{+}\alpha_{+}-\alpha_{-}^T K_{-}\alpha_{-})}{2}.
	\end{equation*}
	
	The proposed method can be generalized to multiclass problems by using different strategies. Here,  one-against-rest $(OAR)$ and one-against-one $(OAO)$ strategies will be examined. For a problem with $c$ classes, $OAR$ method teaches $c$ binary classifiers, so that one class will be classified relative to $c-1$ of other classes and at the end of the class, the test sample will be for the class with the most accurate classifier. On the contrary, $OAO$ method creates all possible binary classifications relative to each class pair. In this case, $c(c-1)/2$ classifiers will be created and the test sample will belong to the class acquiring the most victories.

	\section{Results and evaluations}\label{sec3}
	\subsection{Implementation environment and configuration of the used computer}
	MATLAB and Visual $C++$ were used to implement the method and $MOSEK$ toolbox was used to solve optimization problem. This toolbox provides more accurate answers relative to MATLAB functions for big data set and also solves optimization problems in less time. Since some of the data sets have high size and complexity, part of the tests related to bigger data set were conducted on a system with Windows 7 operating system and 16-core processors and 32 GB of RAM, and the tests related to smaller data set were conducted on a system with a Windows 8 operating system, 8-core processors and a RAM of 8 GB. 
	
	\subsection{Evaluation criteria and evaluation method}
	The main criterion to evaluate the proposed method in this research is the accuracy criterion on test data. The accuracy of a classifier on the test data set equals the ratio between correct predicted data to the total data set of the test.   {In order to have a good estimation of classifiers accuracy, we entered random noise in class and feature level in datasets and used Holdout validation for determining train and test sets.} So, the data set randomly was divided into separate train and test set.  In this case, classifier training and the test is  {done} for $k$ times. In each round, one of the divided parts will be discarded and the classifier will be trained with the rest parts. This process continues for $k$ times. Therefore, in this method, all data will be trained equally and tested one time. Also in this method, the accuracy equals the sum of all data classified correctly divided by all data of initial data set. Since the strong point of the proposed method is on noisy data set, we have tried to use noisy data set more in our tests, but in order to evaluate resistibility and scalability of the proposed method, data set with different dimensions and sizes are used.  {In general, the random noise can be added as follows: compute the random noise and add the noise to the dataset, dataset partition into two-part, training data and test data, and perform a classifier on these data. In evaluating a new method, we generate Gaussian noise and randomly set the data attribute. We also made noise on the data class, which means we randomly changed the data class.}

	\subsection{Introducing used data set and experimental results}
	Owing to the similarity of our proposed method with the spirit of $LMC-HD$, we have used the data set which have already tested by Cevikalp and Triggs (2013) \cite{Cevikalp_2013} as well.
	
	\subsubsection{Volatile Organic Compound identification (VOC) data set}
	This data set\footnote{http://users.rowan.edu/~polikar/RESEARCH/vocdb.html}  has 384 samples and five class, including ethanol, octane, toluene, xylene, trichloroethylene. Each one of the samples has 6 features. In real usage, in fact, we seek to estimate volatile organic compounds. Each of the features was calculated by one of the  6 sensors, and according to the calculations, each one of the samples will be assigned to one of the classes. Proper classification of the data set is important because the sensors calculator of volatile organic compounds are not independent and presence of different elements can influence the estimations of the extent of a certain element and in fact, the data set includes unwanted noisy data. In this data set, using cross-validation, kernel parameter was considered  $\sigma=0.5$ . The results of the tests (table \ref{tab1}) show that the proposed method in sum is more accurate than other methods.

	\begin{table}[]
		\centering
		\caption{Comparing results for $VOC$ data set.}
		\label{tab1}
		\begin{tabular}{|l|l|l|l|l|}
			\hline
			data set& QLMC-HD & LMC-HD &LMC-AH  &SVM  \\ \hline
			VOC &$95.8 \pm 1.1$  &$95.3 \pm 1.1$  &$92.1 \pm 3.6$  & $93.5 \pm 1.7$ \\ \hline
		\end{tabular}
	\end{table}
	
	\subsubsection{Caltech-4 data set}
	This data set\footnote{http://www.vision.caltech.edu/html-files/archive.html}  consists 2876 sample and four classes, including airplanes, cars, faces, motorcycles. Except for the samples of faces class, the samples of three other class were filmed from a side view. The biggest class is airplanes class with 1074 samples and the smallest class is the faces with 450 samples. The images of this dataset have different backgrounds, and sometimes the elements of other classes have emerged in the background with less intensity. Therefore, a collection of the features is used, so that there will be no need for aligning the data. In this data set, using cross-validation method, kernel parameter is considered $\sigma=2$. Our experimental results show that the proposed method is more accurate than former methods. The results are presented in table \ref{tab2}.

	\begin{table}[]
		\centering
		\caption{Comparing results for $Caltech-4$ data set.}
		\label{tab2}
		\begin{tabular}{|l|l|l|l|l|}
			\hline
			data set& QLMC-HD & LMC-HD &LMC-AH  &SVM  \\ \hline
			Caltech-4 &$77.1 \pm 4.4$  &$76.9 \pm 4.3$  &$73.0 \pm 4.3$  & $74.2 \pm 4.8$ \\ \hline
		\end{tabular}
	\end{table}
	
	\subsubsection{UCI repository data set}
	In this section, eight data set were used to compare the proposed method with former methods. The data set were extracted from UCI repository database \footnote{http://archive.ics.uci.edu/ml} and they have fewer dimensions (features) in comparison to other data set. Table \ref{tab3} displays the details of this data set. These data set were tested using $SVM, LMC-AH, LMC- HD,$ and $QLMC-HD$ and Kernel parameter were adjusted for each dataset. We have also used the radial basis function kernel (RBF kernel). The tests results and kernel parameter values for each data set are presented in table \ref{tab4}. The results of the tests indicate that $SVM$ method had a better performance than other methods only in one data set, and $LMC-HD$ method had a better performance in four data set in compare with $SVM$ and $LMC-AH$ methods and had similar results with $LMC-AH$ method in three data set. $QLMC-HD$ method was more accurate in comparison to other methods in four data set. In three cases, since the main parameter $B$ was equal to zero, the final results were the same for $QLMC-HD$ and $LMC-HD$ methods.

	\begin{table}[]
		\centering
		\caption{Characteristics of data set selected from UCI repository}
		\label{tab3}
		\begin{tabular}{|l|l|l|l|}
			\hline
			data set & Number of Classes & Number of samples & Number of feature \\\hline
			Ionosphere & 2 & 351 & 34 \\\hline
			Iris & 3 & 150 & 4 \\\hline
			IS & 7 & 2310 & 19 \\\hline
			LR & 26 & 20000 & 16 \\\hline
			MF & 10 & 2000 & 256 \\\hline
			PID & 2 & 768 & 8 \\\hline
			Wine & 3 & 178 & 13 \\\hline
			WDBC & 2 & 569 & 30\\ \hline
		\end{tabular}
	\end{table}
	
	\begin{table}[]
		\centering
		\caption{Comparing results for UCI repository data set.}
		\label{tab4}
		\begin{tabular}{|l|l|l|l|l|}\hline
			data set & QLMC-HD & LMC-HD & LMC-AH & SVM \\\hline
			Ionosphere $(\sigma=1.5)$  & $94.0\pm 3.6$ & $94.0\pm 3.1$ & $93.7\pm3.4$ & $92.9\pm3.2$ \\\hline
			Iris $(\sigma=2)$ & $96.7\pm 2.3$ & $96.7\pm 2.3$& $94.7\pm 2.9$ & $95.3\pm 3.8$ \\\hline
			IS $(\sigma=0.4)$& $97.3\pm 0.2$ & $97.2\pm 0.3$ & $95.3\pm 0.7$ & $97.1\pm 0.4$  \\\hline
			LR $(\sigma=3)$ &$97.5\pm 0.4$  & $97.5\pm 0.4$ &$97.5\pm 0.4$  &$96.5\pm 0.3$  \\\hline
			MF $(\sigma=4)$ & $98.4\pm 0.4$ &$98.3\pm 0.5$  & $98.3\pm 0.5$ &$98\pm 0.4$ \\\hline
			PID $(\sigma=35)$ &$78.6\pm 1.2$  &$78.4\pm 1.3$  &$77.5\pm 1.9$  & $78.1\pm 1.7$ \\\hline
			Wine $(\sigma=3)$ &$97.8\pm 1.8$  & $98.8\pm 1.6$  & $98.8\pm 1.6$ &$98.2\pm 1.6$ \\\hline
			WDBC $(\sigma=4)$ & $97.1\pm 0.6$ &$97\pm 0.5$  &$96\pm 0.8$  &$97.4\pm 0.9$\\\hline
		\end{tabular}
	\end{table}

	\subsubsection{The noisy data set}
	In this section, given that a strong point of the proposed method is its performance regarding noisy data and outliers, we try to do some tests on such kinds of data set. In general, noise is unwanted and undesirable for each signal. Noises can be generated accidentally or non-accidentally. In fact, the outcome of the measurements equals the noises plus clear data. The practical and real data set are not ideal and they are influenced by many noises and this leads to incorrect data set interpretation. Therefore, the developed method and the decisions taken based on this model will be far from reality. The noises do not include important information and can lead to overfitting. Current research has tried to overcome this problem. The noises can reduce system efficiency in terms of classification, increase classifier’s size and its training time \cite{tan2006introduction}. Almost one of the features of the real data set is that the training set or the test set includes noisy data. The noises usually come up due to human errors during data collection, data transfer, or the constraints of measurement devices. These issues lead to errors in the features (feature noise) or class (class noise) of the data and the noises will occur.

	According to the method mentioned above, different noise was inserted on the class of $ionosphere, iris,$ and $IS$ and the results are presented in the following tables. Table \ref{tab5} shows the results of testing the methods on data set with $\%5$ noise on data classes. According to the results on these three data set, the proposed method in the state of $\%5$ noise on data classes had better results in comparison to other methods. Table \ref{tab6} shows the results of testing this method on data set with $\%10$ noise on data classes. According to the results on these three data set, the proposed method in the state of $\%10$ noise on data classes had better results in comparison with other methods. Table \ref{tab7} shows the results of testing the methods on data set with $\%15$ noise on data classes. According to the results on these three data set, the proposed method in the state of $\%15$ noise on data classes had better results in comparison to other methods. As can be seen, since the noises do not have such a significant impact on the structure of the outliers, an increase in noise on data classes merely leads to a tiny improvement in the results of the proposed method in comparison with the former methods.

	\begin{table}[]
		\centering
		\caption{Results of testing the methods on data set with $\%5$ noise on data classes.}
		\label{tab5}
		\begin{tabular}{|l|l|l|l|l|}
			\hline
			data set &QLMC-HD  & LMC-HD &LMC-AH  & SVM \\ \hline
			ionosphere& $91.7\pm 3.2$ & $91.6\pm 3.1$ &$90.9\pm 3.4$  &$90.4\pm 3.1$  \\ \hline
			Iris & $96.2\pm 2.4$ & $96.1\pm 2.3$ &$93.5\pm 2.2$  & $92.3\pm 3.1$ \\ \hline
			IS& $97.1\pm 0.2$ &$96.7\pm 0.9$  & $94.3\pm 0.8$ & $96\pm 0.9$ \\ \hline
		\end{tabular}
	\end{table}
	
	\begin{table}[]
		\centering
		\caption{Results of testing this methods on data set with $\%10$ noise on data classes.}
		\label{tab6}
		\begin{tabular}{|l|l|l|l|l|}
			\hline
			data set &QLMC-HD  & LMC-HD &LMC-AH  & SVM \\ \hline
			ionosphere& $86.7\pm 3.1$ & $86.3\pm 3.4$ &$85.5\pm 3.2$  &$82.3\pm 3.3$  \\ \hline
			Iris & $90.1\pm 2.6$ & $90\pm 2.3$ &$88.3\pm 3.2$  & $85.2\pm 3.4$ \\ \hline
			IS& $92.7\pm 0.2$ &$91.2\pm 0.6$  & $88.3\pm 0.6$ & $91.4\pm 0.3$ \\ \hline
		\end{tabular}
	\end{table}
	
	\begin{table}[]
		\centering
		\caption{Results of testing the methods on data set with $\%15$ noise on data classes}
		\label{tab7}
		\begin{tabular}{|l|l|l|l|l|}
			\hline
			data set &QLMC-HD  & LMC-HD &LMC-AH  & SVM \\ \hline
			ionosphere& $82.3\pm 3.1$ & $80.4\pm 3.3$ &$79.9\pm 3.1$  &$74.2\pm 3.1$  \\ \hline
			Iris & $84.2\pm 2.3$ & $84.1\pm 2.1$ &$82.2\pm 3.4$  & $79.7\pm 3.3$ \\ \hline
			IS& $86.1\pm 0.3$ &$85.8\pm 0.4$  & $79.6\pm 0.4$ & $81.2\pm 0.6$ \\ \hline
		\end{tabular}
	\end{table}
	
	Now we compare our method with other methods in dealing with data set that the noises have entered in their features, Noisy Train - Noisy Test. According to the above-mentioned method, noises with different percentage were employed on the features of ionosphere, iris, and IS. The noises were applied both regarding test data and train data, to consider the worst state. The Results of the tests on the features of test and training data with $\%5$ noise are presented in table \ref{tab8}. As the results show, in the case of $\%5$ noise on the features of test and training data, the proposed method has improved $\%1$ in compare with $LMC-HD$. 
	
	The results of the tests on data set with $\%10$ noise on the features of test and training data (table \ref{tab9}) show that the proposed method has improved $\%2$ in compare with $LMC-HD$ method. The results of the test on data set with $\%15$ noise on the features of test and training data set are presented in table \ref{tab10}. The results show that in the state of $\%15$ noise on the features of test and training data, the proposed method has improved $\%2$ in comparison to $LMC-HD$ method. According to the results of the comparisons, with an increase in noise percentage, the accuracy of the proposed method declines with a gentler slope, because the additional dimensions will be ignored due to applying hypersphere to the $QSVDD$ method and analyzing affine hull and singular values. Also, the outliers created due to the noises are ignored by the applied $QSVDD$ method.  
	
	\begin{table}[]
		\centering
		\caption{Results of the experimetns on the features of test and training data set with $\%5$ noise.}
		\label{tab8}
		\begin{tabular}{|l|l|l|l|l|}
			\hline
			data set &QLMC-HD  & LMC-HD &LMC-AH  & SVM \\ \hline
			ionosphere& $91.1\pm 4.3$ & $90.6\pm 3.2$ &$89.9\pm 4.2$  &$88.3\pm 3.4$  \\ \hline
			Iris & $91.2\pm 3.7$ & $90.5\pm 2.1$ &$89.5\pm 3.1$  & $89.2\pm 2.1$ \\ \hline
			IS& $95.1\pm 1.3$ &$94.2\pm 1.7$  & $91.3\pm 2.6$ & $90\pm 2.2$ \\ \hline
		\end{tabular}
	\end{table}
	
	\begin{table}[]
		\centering
		\caption{Results of the tests on data set with $\%10$ noise on the features of test and training data}
		\label{tab9}
		\begin{tabular}{|l|l|l|l|l|}
			\hline
			data set &QLMC-HD  & LMC-HD &LMC-AH  & SVM \\ \hline
			ionosphere& $90.2\pm 3.9$ & $88.7\pm 2.9$ &$86.9\pm 2.7$  &$84.3\pm 1.4$  \\ \hline
			Iris & $90.2\pm 2.7$ & $88.5\pm 2.1$ &$86.5\pm 3.1$  & $83\pm 2.1$ \\ \hline
			IS& $94.1\pm 2.7$ &$92.9\pm 1.7$  & $88.3\pm 2.6$ & $86\pm 2.2$ \\ \hline
		\end{tabular}
	\end{table}
	
	\begin{table}[]
		\centering
		\caption{Results of the test on data set with $\%15$ noise on the features of test and training data set.}
		\label{tab10}
		\begin{tabular}{|l|l|l|l|l|}
			\hline
			data set &QLMC-HD  & LMC-HD &LMC-AH  & SVM \\ \hline
			ionosphere& $88.1\pm 2.7$ & $86.6\pm 1.7$ &$83.8\pm 2.4$  &$81.2\pm 3.1$  \\ \hline
			Iris & $88.1\pm 1.6$ & $86.6\pm 2.2$ &$83.4\pm 3.6$  & $79\pm 2.6$ \\ \hline
			IS& $92.3\pm 2.4$ &$90.9\pm 1.9$  & $85.4\pm 2.2$ & $82\pm 2.8$ \\ \hline
		\end{tabular}
	\end{table}
	
	\FloatBarrier

	\section{Conclusion}\label{sec4}
	
	Our proposed method, because of using a geometric hyperdisk model similar to the one used by LMC-HD method, has the advantages of both affine hull and hypersphere geometric models. This method easily can be generalized to multi-class problems and is more accurate than former methods. Due to using affine hull geometric model, it easily ignores unrelated and noisy dimensions and it succeeds more in dealing with outliers.  {So it causes to have the widest margin in comparison with other large margin classifiers.  It can be concluded that QLMC-HD is also more general relative to the LMC-HD method.}
	
	The recent claim is proved by experimental results. The experimental results show that our method at worse conditions shows a behavior like LMC-HD method, and this is because of the more general approach adopted by the proposed method in comparison to the LMC-HD method. With an increase in noise rate of the data set, our method shows better resistance to the noise data, and in comparison to the other methods, an increase in noise data will not lead to a significant reduction in accuracy. Our proposed method in comparison with LMC-HD method in its worst state can lead to equal results. By assuming $B=0$, our proposed method and LMC-HD will be equal. This method can also be upgraded to non-linear issues by using the kernel trick and multi-class issues by combining binary classifiers. So, our method is expected to yield better results in dealing with noises and reduce their impact in determining the center and radius of hyperdisk.
	
	The other idea that will be on our agenda for the future works is to take into consideration the data covariance when making hyperspheres in QSVDD method. In this case, the results at all times will not be equal to the hypersphere and the shape will turn into a hyperellipsoid and we will have a more accurate representation of the data. 
	
	\section*{Acknowledgements}
	Authors wish to thank Omid Abbaszadeh for his useful comments and suggestions which improved the quality of this work.


	\bibliographystyle{abbrvnat}
	\setcitestyle{authoryear}
	\bibliography{library}
\end{document}